\documentclass[11pt]{article}

\usepackage[preprint]{acl}

\usepackage{times}
\usepackage{latexsym}

\usepackage[T1]{fontenc}

\usepackage[utf8]{inputenc}

\usepackage{microtype}

\usepackage{inconsolata}

\usepackage{graphicx}

\usepackage{tabularx}
\usepackage{booktabs}   
\usepackage{array}      
\usepackage{tabularx,makecell} 
\usepackage{fancyhdr}
\title{Single layer tiny $Co^{4}$ outpaces GPT-2 and GPT-BERT}

\author{Noor Ul Zain\textsuperscript{1}, 
    \textbf{Mohsin Raza}\textsuperscript{1}, \textbf{Ahsan Adeel}\textsuperscript{1,*} \\
    \textsuperscript{1} CMI-Lab \\  University of Stirling, UK \\
    \textsuperscript{*} \texttt{ahsan.adeel1@stir.ac.uk}}

\begin{document}
\maketitle
\begin{abstract}
We show that a tiny $Co^{4}$ machine \cite{adeel2025beyond} with a single layer, two heads, and 8M parameters, operating at an approximate cost of $O(N)$ (where \textit{N} is the number of input tokens), outpaces the BabyLM Challenge baselines GPT-2\footnote{https://huggingface.co/BabyLM-community/babylm-baseline-10m-gpt2} (124M, 12 layers, $O(N^2)$) and GPT-BERT\footnote{https://huggingface.co/BabyLM-community/babylm-baseline-10m-gpt-bert-causal-focus} (30M, 12 layers, $O(N^2)$) in just two epochs, while both are trained for ten. $Co^{4}$ achieves orders-of-magnitude greater training efficiency on 10M tokens, demonstrating highly sample-efficient pretraining. Using the BabyLM challenge evaluation pipeline across complex benchmarks, $Co^{4}$ exhibits strong zero-shot and fine-tuning performance on SuperGLUE tasks. Specifically, $Co^{4}$ outperforms GPT-2 on 5 out of 7 zero-shot metrics and 6 out of 7 fine-tuning tasks, and GPT-BERT on 4 out of 7 metrics in both cases. These results suggest the need to rethink prevailing deep learning paradigms and associated scaling laws.
\end{abstract}

Cellular neurobiological evidence \cite{suzuki2023deep, marvan2024cellular} on how mammalian brains achieve fast and flexible computation continues to challenge deep (hierarchical) learning \cite{lecun2015deep, vaswani2017attention, wang2025hierarchical}, predictive coding \cite{rao1999predictive, friston2005theory, friston2010free}, and scaling laws \cite{kaplan2020scaling}. Evidence suggests that the brain’s computational power lies in shallow architectures, where cortical and subcortical networks operate with massive parallelism, leveraging cortical microcircuits and thalamo-cortical loops \cite{aru2020cellular, storm2024integrative, Phillips2024cellular} to support faster, context-sensitive, and coherent internal understanding \cite{adeel2025beyond}.\\
Modern deep learning architectures, such as Transformers \cite{vaswani2017attention, jaegle2021perceiver, alayrac2022flamingo}, which underpin models like GPT and GPT-BERT, act as sequential local agents reducing predictive error or free energy \cite{friston2005theory, friston2010free}, yet without regard for local coherence \cite{marvan2024cellular}. During the feedforward (FF) phase, they lack intrinsic mechanisms to judge the true relevance of an attended token \cite{adeel2025beyond}. Instead, relevance is indirectly shaped by backpropagation during the feedback (FB) phase, a brute-force, reward-driven process. Incoherent inferences generated by initial agents (e.g., early transformer blocks) propagate to subsequent agents, where they are reinforced through ineffective FB signals. We refer to this as a "Chinese Whispers" problem.\\
Consequently, these deep nets require vast datasets, extensive training time, and significant compute, resulting in unsustainable economic, environmental, and technical costs \cite{thompson2020computational}. The reliance on deeper architectures for hierarchical feature abstraction is a shared limitation across other neural models, including long short-term memory (LSTM) \cite{6795963}, gated-recurrent units (GRUs) \cite{chung2014empirical}, and convolution neural networks (CNNs) \cite{lecun1989backpropagation}.\\
The recently proposed $Co^4$ machine \cite{adeel2025beyond} emulates higher-level perceptual processing (HLPP) and awake thought (AT) mental states \cite{Phillips2024cellular}. Within a single layer, during FF, it executes triadic FB loops among latent questions (Qs), clues (Ks), and hypotheses (Vs), enabled by three two-point neurons (TPNs)\footnote{A pyramidal two-point neuron in the mammalian neocortex integrates feedforward input at its basal site and contextual input at its apical dendrites. When both are aligned in time, the neuron fires bursts that amplify coherent, contextually relevant signals for active inference.} \cite{aru2020cellular, storm2024integrative, Phillips2024cellular}, each representing an agent holding K, Q, and V. Unlike Transformers, which propagate layer-wise, $Co^4$ enables all agents to co-evolve Qs, Ks, and Vs in parallel: Qs update based on Ks and Vs; Ks update based on Qs and Vs; Vs evolve based on Ks and Qs. Each TPN agent independently forms distinctive Q–K–V perspectives, thereby maximizing local and global coherence \cite{marvan2024cellular} while minimizing free energy \cite{friston2005theory, friston2010free}, ensuring token relevance before attention is applied or decisions are made. This cooperative mechanism enables diverse, parallel, and deep reasoning chains without requiring additional layers, at an approximate cost of $O(N)$ \cite{adeel2025beyond}.\\
This paper is the first to report the $Co^4$ machine's performance on complex language benchmarks. From a cognitive modeling perspective, we compare training trajectories of $Co^4$, GPT-2, and GPT-BERT to those of children using psycholinguistic metrics under data-limited conditions modeled after human language acquisition \cite{charpentier2025babylm}. Despite its tiny size, just one layer, two heads, and 8M parameters, $Co^4$ (with $O(N)$ cost) outpaces GPT-2 (124M parameters) and GPT-BERT (30M), both using 12 layers ($O(N^2)$ cost), achieving orders-of-magnitude greater efficiency and stronger generalization on a 10M-token dataset.
\section{Neurons and $Co^4$ agents with two points of input integration}
Going beyond the $20^{th}$-century neuroscience conception of point neurons (PNs) \cite{hausser2001synaptic}, on which most current brain theories and AI systems are based, $21^{st}$-century neuroscience \cite{larkum1999new, phillips2017cognitive, phillips2023cooperative, larkum2013cellular, major2013active, ramaswamy2015anatomy, larkum2022dendrites, adeel2020conscious, kording2000learning, SchumanAnnual, poirazi2020, larkum2018perspective, shine2016dynamics, shine2019human, shine2019neuromodulatory, shine2021computational, schulz2021gaba, kay2020contextual, kay2022comparison} has revealed that certain neurons, particularly some pyramidal neurons in the mammalian neocortex, integrate inputs at two distinct locations. These are often referred to as TPNs, which combine information from the external environment (feedforward (FF)) at one site (basal) and contextual (C) input at another (apical). TPNs trigger high-frequency firing (bursting) when the FF and C inputs are matched in time, that is,
when both the basal and apical zones are depolarized. This results in the amplification of coherent signals, enabling enhanced contextually rich processing \cite{Phillips2024cellular}. \\
The flexible interaction between FF and C inputs is suggested to be the hallmark of conscious processing \cite{aru2020cellular, storm2024integrative, marvan2021apical} and linked to distinct mental states, including wakefulness (WF), slow-wave (SW)
sleep, and rapid eye movement (REM) sleep \cite{Phillips2024cellular}. Dysfunctional interactions between FF and C inputs have been linked to intellectual learning disabilities \cite{nelson2021dendritic, granato2024dysfunctions}.\\
Several TPN-inspired machine learning algorithms have been proposed to flexibly combine top-down C and bottom-up FF information streams \cite{payeur2021burst, Greedysingle, guerguiev2017towards, sacramento2018dendritic, IllingLocal, Greedysingle, zenke2017continual, kirkpatrick2017overcoming, kastellakis2016linking, bono2017modeling, limbacher2020emergence}. However, most of these efforts have focused on using apical (contextual) inputs primarily for learning. Ample evidence suggests that the apical site not only receives feedback from higher perceptual levels but also integrates simultaneous events across multiple hierarchical levels while processing FF information. For example, results using TPN-inspired CNNs \cite{adeel2020conscious, adeel2022context, adeel2022unlocking, raza2024overlooked} showed that these architectures could drastically reduce the transmission of conflicting FF signals to higher perceptual areas, achieving orders-of-magnitude reductions in the number of neurons needed to process heterogeneous real-world audio-visual data, compared to standard PN-based CNNs.\\
More recent findings demonstrate that the TPN-inspired $Co^{4}$ machine \cite{adeel2025beyond}, emulating higher level perceptual processing and imaginative thought mental states can enable significantly faster learning with substantially lower computational demands (e.g., fewer heads, layers, and tokens) at an approximate cost of $O(N)$. These gains were observed across a variety of domains, including reinforcement learning, computer vision, and natural language question answering.\\
These efforts to develop efficient machine learning models align with scaled-down pretraining using fewer than 100M tokens, evaluating language models (LMs) on the same types and quantities of data that humans are exposed to \cite{charpentier2025babylm}. The aim is to build plausible cognitive models of human learning and to better understand how children are exposed to language with such efficiency. By combining cellular neurobiologically inspired, TPN-based $Co^{4}$ machine \cite{adeel2025beyond} with this scaled-down pretraining strategy, we introduce the $Co^{4}$ LM.

\begin{figure*}[t]
  \includegraphics[width=0.3\linewidth]{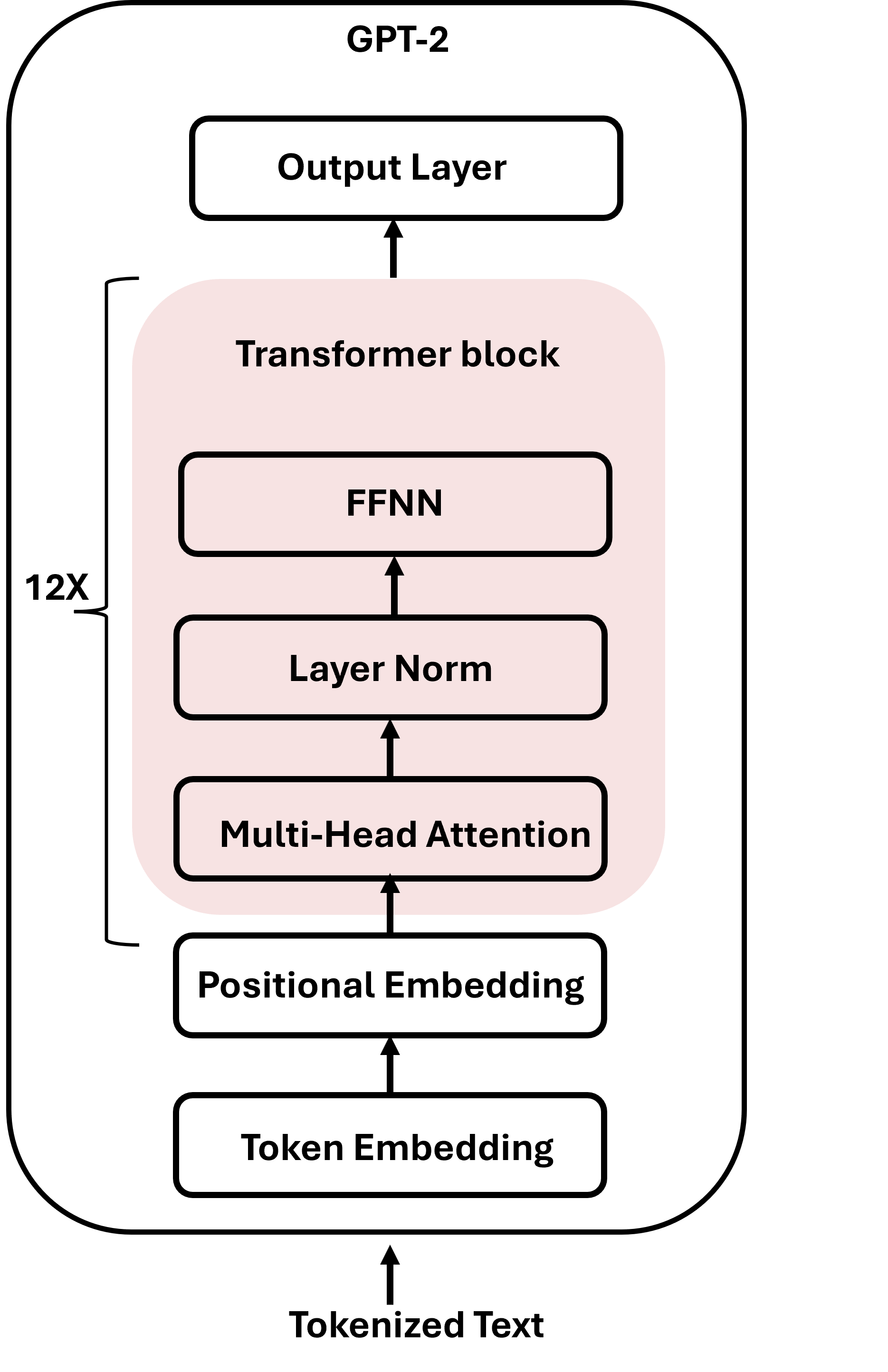} \hfill
  \includegraphics[width=0.68\linewidth]{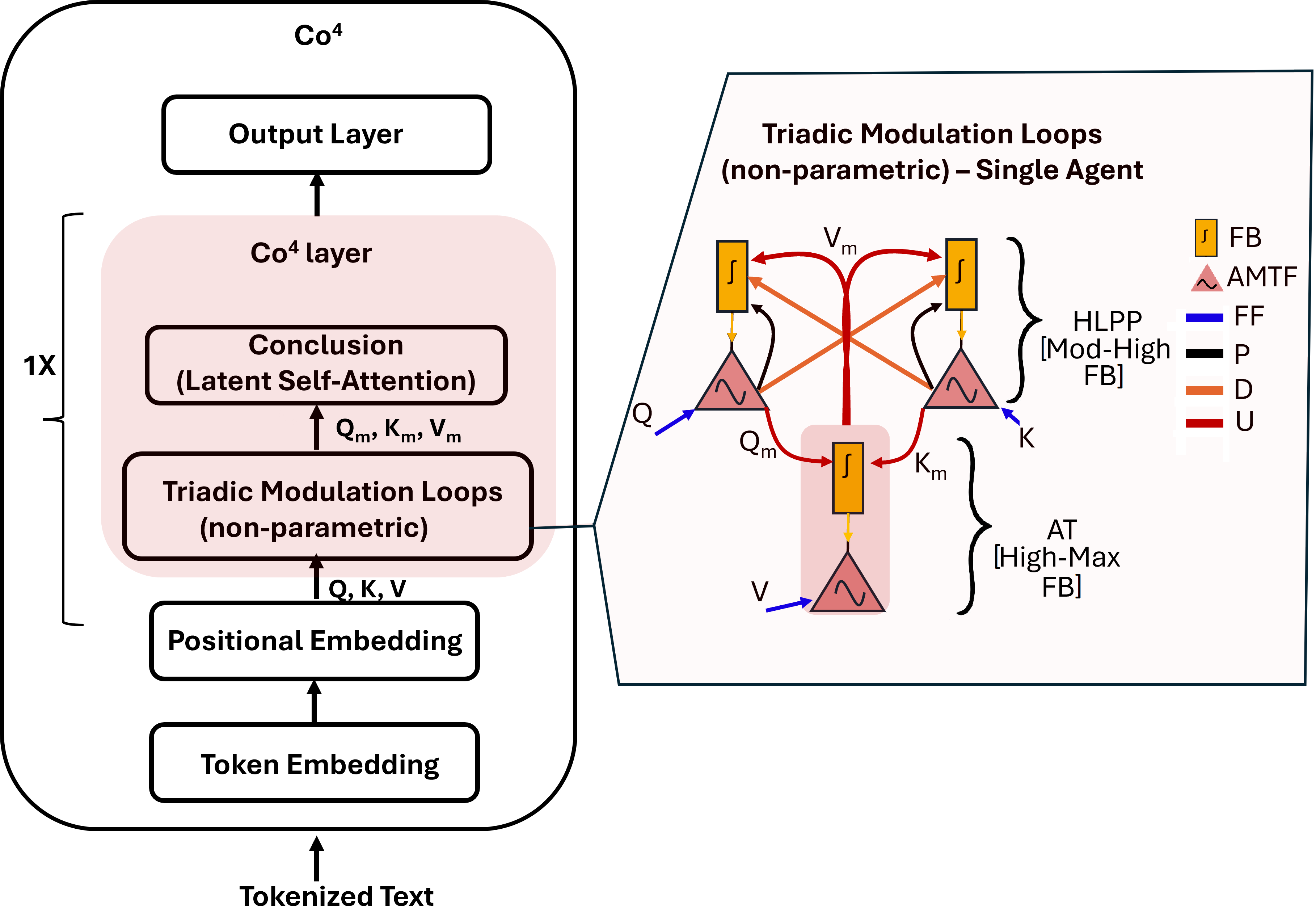}
  \caption [GPT-2 vs. Co$^{4}$ overview]{Language Models: GPT-2 (Left) vs. $Co^{4}$ (Right). In $Co^{4}$, the learnable parameters are only in the embedding layer and the initial Q, K, V representations, followed by a single layer of non-parametric triadic modulation loops (referred to as “1x” Co4 or single-layered Co4). $Co^{4}$ does not require feed feed-forward neural network (FFNN/ MLP) layer used in standard GPT-type architectures. Inside these loops, three populations of three pyramidal two-point processors, each associated with Q, K, and V, respectively, simultaneously integrate FF information and FB context at two functionally distinct sites. The apical (top-down) site (shown in the rectangle) integrates context, while FF information is integrated at the basal (bottom-up) site (shown in the triangle). Each processor, via asynchronous modulation (MOD) transfer functions\footnotemark, operating in higher-level perceptual processing (HLPP) or awake thought (AT) mode, depending on the strength of FB, amplifies FF transmission if it is relevant in that context (represented by P, D, U). Otherwise, it attenuates the signal, resulting in the selective amplification of coherent FF information  \cite{adeel2025beyond}. P, D, and U, along with the credit assignment (reward) coming from the higher perceptual layer (teacher), can be seen as dynamic local competitive normalization and global cooperative organisation, respectively. This ensures that local and global coherence and consistency are maximized \cite{marvan2024cellular}, while prediction error or free energy \cite{friston2005theory, friston2010free} is minimized, enabling a deeper form of "real understanding". A combination of three TPNs and one loop constitutes one agent. A set of 12 agents with 12 loops runs in parallel, evolving their Qs, Ks, and Vs simultaneously, before applying latent self-attention at  $O(L \times N)$ where \textit{L} is a small fraction of the input sequence length, making the overall cost approximately $O(N)$.}
  
\end{figure*}

\section{$Co^4$ Language Model} 
Figure 1 (left) illustrates the standard GPT-2 model, consisting of 12 Transformer layers, where each layer performs a simple conclusion via self-attention ($QK^TV$) at the cost of $O(N^2)$. This can be interpreted as 12 agents working sequentially. The selection of relevant and irrelevant tokens in the FF phase is determined through backpropagation, a brute-force process solely driven by the global objective. This rigidity causes the network to depend heavily on pre-learned patterns, limiting its ability to generate new perspectives quickly. When initial thoughts are misleading, arriving at a correct conclusion may require significantly more time and computation, or may not happen at all, due to limited internal flexibility and constrained cognitive resources \cite{adeel2025beyond}.\\
In contrast, Figure 1 (right) shows a single-layer $Co^4$ machine with two attention heads. After initializing the latent queries (Qs) as a set of neuronal agents (e.g., 24) (as opposed to 12 attention blocks + feedforward neuron network (FFNN) in GPT-2 and GPT-BERT), they begins to co-evolve their own Qs, Ks, and Vs in parallel during the FF phase via triadic modulation loops leveraging proximal (P), distal (D), and universal (U) contextual fields. This co-evolution is enabled through inherent, moment-by-moment cooperation mechanisms or asynchronous modulation (MOD) transfer function \cite{adeel2025beyond}, resulting in rich, contextually-aware, and diverse parallel reasoning chains at the cellular level. Each agent independently develops its own Q, K, and V, leading to 24 attention maps and 24 possibly different conclusions. Importantly, this all occurs virtually, allowing the model to pre-select relevant tokens before applying latent self-attention at an approximate cost of $O(N)$ \cite{adeel2025beyond}.\\
The $Co^4$ language model frames text generation as an autoregressive, left-to-right process: given a prefix of tokens, the model computes a probability distribution over the next token via a softmax applied to its hidden state. We use the same tokenizer as the baselines. The input tokens are first mapped to continuous vectors through a embedding layer and are augmented with positional embeddings to encode sequence order. During training, a triangular causal mask ensures that each position can only attend to previous positions. The model’s weights are optimized by minimizing the cross-entropy (CE) loss (equivalently, the negative log-likelihood) of the true next token. \\
The $Co^4$ language model condenses this pipeline into a single decoder layer with just two attention heads, yet enriches it via triadic modulation loops among Q-, K-, and V-TPNs, operating through P, D, and U contextual fields \cite{adeel2025beyond, adeel2020conscious}. After token embedding and positional projection, each token’s Q, K, and V vectors co-evolve through a series of rapid and modulated updates.\\
We trained $Co^4$ on a 10M-token slice of the BabyLM corpus \cite{babylm_gpt2_baseline}, using the same autoregressive CE objective but at a fraction of the training budget of GPT-2 and GPT-BERT, which are the official baselines provided by the organizers of this challenge. More details related to the hyperparameters for these baselines can be found on the relevant model repositories on Hugging Face.

\section{Results}
In this section, we present the performance of our tiny $Co^4$ machine across a range of language modeling benchmarks. The seven tasks described first assess the model’s linguistic capabilities in a purely zero-shot setting, without any additional training or fine-tuning. Later in the section, we also evaluate $Co^4$'s performance on fine-tuning benchmarks and provide an extensive comparison with the baseline.
\\ We utilize the evaluation suite from the BabyLM Challenge \cite{charpentier2025babylm}, which includes the following zero-shot metrics. The first two, newly introduced, are designed to compare the language model's responses to those of human judgments and behavioral data.

\begin{itemize}
    \item Eye Tracking and Self-paced Reading: This psycholinguistic measure evaluates whether the model can mimic the eye tracking and reading time of a human by using the surprisal of a word as a proxy for time spent reading a word \cite{de2024cloze}.
    \item WUGs: morphological Adapting the classic “Wug” paradigm, this evaluates whether models can generalize morphological rules to form novel noun derivatives from unseen adjectives, and compares the model's generalization to that of humans \cite{Hofmann2025}.
    \item Entity Tracking: Probes a model’s capacity to update and maintain the state of entities throughout a narrative or dialogue by asking it to predict an entity’s final condition after a series of changes \cite{kim2023entity}.
    \item EWoK: This benchmark evaluates the model's internal world knowledge across domains like spatial relations and social interactions \cite{ivanova2024elements}.
    \item BLiMP: Testing various grammatical phenomenon, the Benchmark of Linguistic Minimal Pairs evaluates whether a model consistently picks the grammatically correct alternative from a pair of minimally different sentences \cite{warstadt2020blimp}.
    \item BLiMP Supplement: This is a supplement to BLiMP and was introduced in the first edition of the BabyLM challenge. It is more focused on dialogue and questions \cite{warstadt2025findings}.
\end{itemize} 

\footnotetext{For the mathematical details of these functions and the core mechanism behind triadic modulation loops, please check \cite{graham2025context}.}

The metrics used to evaluate the model on each of these zero-shot benchmarks are as follows:
\begin{itemize}
\item Accuracy in predicting the correct completion or sentence for BLiMP, BLiMP Supplement, EWoK, Entity Tracking, and WUGs.
\item Change in $R^2$ prediction from baseline for Eye Tracking and Self-paced Reading.
\end{itemize} 

Table \ref{t.metrics_3} shows the performance of tiny $Co^{4}$ language model on the metrics outlined above. As shown, our computationally efficient model, $Co^4$-$\alpha$, outperforms GPT-2 on 5 out of 7 metrics. As for GPT-BERT, another configuration $Co^4$-$\beta$, outperforms it on 4 out of 7 metrics. 
These hyperparameters for these configurations are further outlined in the Appendix.

\begin{table}[ht]
  \centering
  \begin{tabularx}{\columnwidth}{lXXXX}
    \hline
    \textbf{Metric} & \textbf{GPT-2} & \textbf{$Co^{4}$-$\alpha$} & \textbf{GPT-BERT} & \textbf{$Co^{4}$-$\beta$} \\
    \hline
    Eye Tracking       & 8.66  & \textbf{8.67}  & \textbf{9.89} & 8.19 \\
    Self-paced Reading & 4.34  & \textbf{4.59}  & 3.45          & \textbf{3.62} \\
    WUGs               & 52.50 & \textbf{68.00} & 43.00         & \textbf{93.00} \\
    Entity Tracking    & 13.90 & \textbf{26.71} & 33.96         & \textbf{41.36} \\
    EWoK               & 49.90 & \textbf{50.01} & 49.49         & \textbf{50.11} \\
    BLiMP              & \textbf{66.36} & 53.55 & \textbf{71.66} & 51.20 \\
    BLiMP Supplement   & \textbf{57.07} & 52.59 & \textbf{63.21} & 49.82 \\
    \hline
  \end{tabularx}
  \caption{\textbf{Zero-shot metrics comparison:} GPT-2 vs. $Co^4$-$\alpha$ and GPT-BERT (causal-focus) vs $Co^4$-$\beta$
The single-layer, tiny $Co^4$ model outperformed GPT-2 on 5 out of 7 metrics, and GPT-BERT on 4 out of 7 metrics, despite being trained at a fraction of the computational cost, \textbf{in 2 epochs}.}
  \label{t.metrics_3}
\end{table}


\begin{table}[ht]
  \centering
  \begin{tabularx}{\columnwidth}{>{\raggedright\arraybackslash}l *3{>{\centering\arraybackslash}X}}
    \hline
    \textbf{Metric} & \textbf{GPT-2} & \textbf{GPT-BERT} & \textbf{$Co^{4}$-$\gamma$} \\
    \hline
    Hypernym                 & 48.93 & 49.05 & \textbf{54.75} \\
    QA Cong. Easy            & 50.00 & 67.19 & \textbf{87.50} \\
    QA Cong. Tricky          & 39.39 & 50.30 & \textbf{53.94} \\
    \makecell[l]{Subject Aux\\Inversion} & \textbf{81.33} & 81.28 & 65.48 \\
    Turn Taking              & 65.71 & \textbf{68.21} & 50.36 \\
    Overall                  & 57.07 & \textbf{63.21} & 62.40 \\
    \hline
  \end{tabularx}
  \caption{\textbf{BLiMP Supplement benchmark}: $Co^4$-$\gamma$ demonstrates superior performance in the BLiMP Supplement benchmark and the individual tasks in this benchmark. Although this configuration of $Co^4$-$\gamma$ does not outperform the psycholinguistic metrics, it outperforms the baselines in the BLiMP Supplement.}
  \label{t.metrics_3_new}
\end{table}

\begin{table}[ht]
  \centering
  \begin{tabularx}{\columnwidth}{l l X X X}
    \hline
    \textbf{Task} & \textbf{Metric} & \textbf{GPT-2} & \textbf{GPT-BERT} & \textbf{$Co^{4}$} \\
    \hline
    MRPC     & F1       & 80.77   & 83.44 & \textbf{84.15} \\
    QQP      & F1       & 62.45   & \textbf{72.03} & 62.73 \\
    BoolQ    & Accuracy & 66.91   & 68.07 & \textbf{69.05} \\
    MNLI     & Accuracy & \textbf{51.12} & 46.86 & 44.25 \\
    MultiRC  & Accuracy & 65.72   & \textbf{68.28} & 66.01 \\
    RTE      & Accuracy & 56.83   & 56.12 & \textbf{59.71} \\
    WSC      & Accuracy & 61.54   & 65.38 & \textbf{67.31} \\
    \hline
  \end{tabularx}
  \caption{SuperGLUE tasks}
  \label{tab:model-comparison}
\end{table}

 Table \ref{t.metrics_3_new} reports performance  of $Co^4$-$\gamma$ on the BLiMP Supplement benchmark. This $Co^4$-$\gamma$ is a different configuration of our architecture, which notably performed better on BliMP Supplement. Since it did not beat most of the metrics, we did not pick it as our best configuration but we wanted to include its superior performance on BLiMP. It should be noted that our model performs better on BLiMP Supplement compared to BLiMP, suggesting that the $Co^4$ model has an inherent bias toward more complex tasks and long-term dependencies characteristic of BLiMP Supplement’s subtasks. More challenging than the original BLiMP benchmark, BLiMP Supplement was introduced in the most recent version of the BabyLM Challenge \cite{charpentier2025babylm}. It is more challenging since models perform relatively lower on it as compared to BLiMP \cite{warstadt2025findings}, and also because it consists of more dialogues and questions as compared to the minimally different sentences in BLiMP. It is comprised of the following five subtasks:
\begin{itemize}
\item Hypernym: Checks whether a word is correctly recognized as a superset or subset of another (e.g., a dog is a mammal, so having a dog implies having a mammal).
\item QA Congruence Easy: Verifies whether the question type matches the answer (e.g., a who question is answered with a person rather than a thing).
\item QA Congruence Tricky: Similar to QA Congruence Easy but with more ambiguous cases.
\item Subject–Aux Inversion: Checks whether the auxiliary verb is correctly inverted with the subject (e.g., Is she coming?).
\item Turn Taking: Checks whether the correct personal pronoun is used when answering a question in dialogue.
\end{itemize}

\textbf{Finetuning:} Table \ref{tab:model-comparison} reports performance on SuperGLUE tasks as part of fine-tuning. \cite{superglue}.  We picked our best $Co^{4}$ configuration overall ($Co^4$-$\alpha$) for the finetuning. Our novel architecture achieves comparable results across most fine-tuning tasks and demonstrates better performance on 6 out of the 7 tasks when compared to GPT-2 and 4 out of the 7 tasks when compared to GPT-BERT. 
These tasks are: 

\begin{itemize}
\item BoolQ: A yes/no question-answering dataset with unprompted and unconstrained questions \cite{clark2019boolq}
\item MNLI: The Multi-Genre Natural Language Inference corpus tests whether a model can recognize textual entailment \cite{williams2017broad}.
\item MRPC: The Microsoft Research Paraphrase Corpus contains pairs of sentences that are either paraphrases (semantically equivalent) or unrelated \cite{dolan2005automatically}.
\item QQP: Similarly to MRPC, the Quora Question Pairs corpus tests a model’s ability to determine whether pairs of questions are semantically similar. These questions are sourced from Quora \cite{babylm_gpt2_baseline}.
\item MultiRC: The Multi-Sentence Reading Comprehension corpus evaluates a model’s ability to select the correct answer from a list of candidates given a question and a context paragraph. In this version, the data is reformulated as a binary classification task judging whether an answer to a question-context pair is correct \cite{khashabi2018looking}.
\item RTE: Recognizing Textual Entailment tests the model’s ability to recognize textual entailment \cite{dagan2005pascal, dagan2022recognizing,  bentivogli2009fifth}.
\item WSC: The Winograd Schema Challenge evaluates coreference resolution in sentences containing a pronoun and a list of noun phrases. This version reformulates the task as a binary classification problem using examples consisting of a pronoun and a noun phrase \cite{levesque2012winograd}.
\end{itemize}
The hyperparameters for this task are outlined in the Appendix. 

\section{Conclusion}
The $Co^4$ model has a computational complexity of $O(L \cdot N + \alpha)$, scaling linearly with the number of input tokens ($N$), where $L$ is the number of latent queries and $\alpha$ is a small fixed overhead. In contrast, models like GPT-2 and GPT-BERT scale quadratically at $O(N^2)$, making them significantly more expensive as input size grows. In standard Transformers, multiply–accumulate (MAC) operations grow with the quadratic term $P^2 \cdot E$ due to self-attention, where $P$ is the number of tokens and $E$ is the embedding dimension. In $Co^4$, this is replaced by a more efficient linear term $L_q \cdot P \cdot E$, enabled by a small set of latent queries. As a result, $Co^4$ achieves substantial computational savings and superior scalability over conventional Transformers.
\\
Despite being a single-layer model, the tiny $Co^4$ machine outperforms GPT-2 and GPT-BERT on most evaluated performance metrics, while requiring only a fraction of the computational resources. \\
Future directions include scaling to larger datasets, integrating multi-objective or hybrid cost functions (e.g., those used in GPT-BERT), and evaluating different modes of apical operation \cite{Phillips2024cellular, graham2024transfer, pastorelli2023two}. In addition, scaling beyond 8M parameters is part of ongoing work.


\section{Acknowledgments}
Advanced Research + Invention Agency (ARIA): Nature Computes Better Opportunity seeds. Professor Bill Phillips, Professor Leslie Smith, Professor Bruce Graham, and Dr Burcu Can Buglalilar from the University of Stirling. Professor
Panayiota Poirazi from IMBB-FORTH, Professor Peter Konig from the University Osnabruck. Professor Heiko Neumann from Ulm University, Dr James Kay from the University of Glasgow, and several other eminent scholars for their help and support in several different ways, including reviewing this work, appreciation, and encouragement. We also acknowledge ChatGPT for its assistance with proofreading.\\
\\
\textbf{Competing interests} 
The authors declare no conflict of interest.

\bibliography{latex/acl_latex}

\clearpage
\appendix
\section{Pre-Training Details}

\begin{table}[h!]
\centering
\small
\setlength{\tabcolsep}{3pt} 
\renewcommand{\arraystretch}{1.1} 
\begin{tabularx}{\columnwidth}{lXXX}
\toprule
\textbf{Hyperparameter} & \textbf{$Co^{4}$-$\alpha$} & \textbf{$Co^{4}$-$\beta$} & \textbf{$Co^{4}$-$\gamma$} \\
\midrule
Number of parameters & 8M & 8M & 8M \\
Number of layers\textsuperscript{$\dagger$} & 1 & 1 & 1 \\
Embedding size & 256 & 256 & 256 \\
Vocabulary size & 16384 & 16384 & 16384 \\
Attention heads & 2 & 2 & 2 \\
Hidden dropout & 0.1 & 0.1 & 0.1 \\
Batch size & 32 & 64 & 32 \\
Sequence length & 512 & 512 & 512 \\
Warmup ratio & 1.3\% & 1.4\% & 1\% \\
Learning rate & 0.0002 & 0.00001 & 0.0002 \\
Learning rate scheduler & constant & constant & cosine \\
Optimizer & ADAMW & ADAMW & ADAMW \\
ADAMW $\epsilon$ & 1e-8 & 1e-8 & 1e-8 \\
ADAMW $\beta_1$ & 0.9 & 0.9 & 0.9 \\
ADAMW $\beta_2$ & 0.999 & 0.999 & 0.999 \\
\bottomrule
\end{tabularx}
\caption{Pre-training hyperparameters for the STRICT-SMALL track across three configurations. \textsuperscript{$\dagger$}One layer refers to a module composed of our custom Co4 layer.}
\label{tab:pretrain_small}
\end{table}


The training procedure, which has been briefly highlighted before, is as follows. We use the same tokenizer as the baselines, with a vocab size of 16384 and a small 1-layer model with the hyperparameters mentioned above. The $Co^4$ language model with a single decoder layer and just two attention heads is trained on the 10M corpus. It is powered via the aforementioned triadic modulation loops among Q-, K-, and V-TPNs, operating through P, D, and U contextual fields. After token embedding and positional projection, each token’s Q, K, and V vectors co-evolve through a series of rapid and modulated updates. 

The main goal was to keep the model as minimal as possible, to see the true power of the biologically-inspired triadic modulation loops within the layer. It is observed that the model performance converges over just a few epochs, i.e., 2 in this case.

\section{Finetuning Details}

We perform a grid search for the following hyperparameters:
\begin{itemize}
    \item \textbf{Number of epochs:} \{3, 5, 10\}
    \item \textbf{Learning rate:} \{$3\times10^{-5}$, $5\times10^{-5}$, $1\times10^{-4}$, $2\times10^{-4}$, $3\times10^{-4}$, $5\times10^{-5}$,
    $5\times10^{-5}$\}
    \item \textbf{Batch size:} \{16, 32, 64\}
\end{itemize}
For WSC (low training data), we expand the search to:
\begin{itemize}
    \item \textbf{Number of epochs:} \{3, 5, 10, 15, 20, 25, 30, 100\}
    \item \textbf{Learning rate:} \{$3\times10^{-5}$, $5\times10^{-5}$, $7\times10^{-5}$, $1\times10^{-4}$, $2\times10^{-4}$, $3\times10^{-4}$, $5\times10^{-4}$\}
    \item \textbf{Batch size:} \{16, 32, 64\}
     \}
\end{itemize}

\end{document}